%% file: paper.tex
\documentclass[runningheads]{llncs}

\usepackage{adjustbox}
\usepackage[shortend,ruled,linesnumbered]{algorithm2e}
\usepackage{amsmath}
\usepackage{amsfonts,amssymb}
\usepackage{array}
\usepackage[english]{babel}
\usepackage{booktabs}
\usepackage{comment}
\usepackage{enumitem}
\usepackage{graphicx}
\usepackage[utf8]{inputenc}
\usepackage{multirow}
\usepackage{setspace}
\usepackage{stmaryrd}
\usepackage{subcaption} 
\usepackage{siunitx}
\usepackage{tablefootnote}
\usepackage{tikz}
\usetikzlibrary{arrows,decorations,fit,positioning,shapes}
\usepackage{empheq}
\usepackage{xcolor}
\usepackage{xspace}
\usepackage{caption}
\usepackage{setspace}

\usepackage{setspace}
\usepackage{enumitem}
\setlist{nolistsep}
\usepackage{listings}

\usepackage{booktabs}   
\usepackage{multirow}   
\usepackage{cite}       
\usepackage{subcaption} 
\usepackage{caption}
\usepackage{todonotes}  
\usepackage{tablefootnote}
\usepackage[most]{tcolorbox}

\let\llncssubparagraph\subparagraph
\let\subparagraph\paragraph
\usepackage{titlesec}
\let\subparagraph\llncssubparagraph

\graphicspath{{figures/}}
\DeclareGraphicsExtensions{.pdf}

\fontfamily{ptm}

\input{defs}

\usepackage[pdftex%
,colorlinks=true       
,linkcolor=midblue        
,citecolor=midblue        
,filecolor=black       
,urlcolor=midblue         
,linktoc=page           
,plainpages=false]{hyperref}
\addto\extrasenglish{%

}

\begin{document}

%
\title{On Validating, Repairing and Refining\\Heuristic ML Explanations}

\author{%
  Alexey Ignatiev\inst{1,3}   \and
  Nina Narodytska\inst{2}   \and
  Joao Marques-Silva\inst{1}
}

\authorrunning{Ignatiev et al.}

\institute{%
  Faculty of Science, University of Lisbon, Portugal
  \{\mailtodomain{aignatiev}\texttt{,}\mailtodomain{jpms}\}\texttt{@ciencias.ulisboa.pt} \and
  VMware Research, CA, USA
  \href{mailto:nnarodytska@vmware.com}{\texttt{nnarodytska@vmware.com}} \and
  ISDCT SB RAS, Irkutsk, Russia
}




\maketitle


\input{abs}

\input{intro}
\input{prelim}

\input{enc}
\input{cases}
\input{res}
\input{relw}
\input{conc}

\clearpage
\bibliographystyle{splncs04}
\bibliography{paper}

\end{document}

%% file: defs.tex
\setlist{nolistsep}
\setitemize{noitemsep,topsep=3pt,parsep=2pt,partopsep=2pt}
\setenumerate{noitemsep,topsep=3pt,parsep=2pt,partopsep=2pt}

\titlespacing{\section}{0pt}{*2.25}{*1.0}
\titlespacing{\subsection}{0pt}{*1.0}{*0.75}
\titlespacing{\subsubsection}{0pt}{*0.75}{*0.5}

\makeatletter
\renewcommand\paragraph{\@startsection{paragraph}{4}{\z@}%
                                    {1.0ex \@plus0.35ex \@minus.15ex}%
                                    {-1em}%
                                    {\normalfont\normalsize\bfseries}}

\makeatother

\makeatletter
\def\thm@space@setup{%
  \thm@preskip=1.5pt \thm@postskip=1.5pt
}
\makeatother



\titleformat{\paragraph}[runin]
            {\normalfont\normalsize\bfseries}{\theparagraph}{0.4em}{}

\newcommand{\fml}[1]{\ensuremath\mathcal{#1}}

\newcommand{\tn}[1]{\textnormal{#1}}
\newcommand{\tl}[1]{\textsl{#1}}
\newcommand{\lit}[1]{\ensuremath\tl{#1}\xspace}
\newcommand{\nlit}[1]{\ensuremath\neg\tl{#1}\xspace}
\newcommand{\eqlit}[2]{(\ensuremath\tl{#1}=\tl{#2})\xspace}
\newcommand{\noarg}{\phantom{~}}

\newcommand{\nhlit}[1]{\ensuremath\textcolor{blue}{\neg\tl{\bfseries{#1}}}\xspace}
\newcommand{\model}[1]{\mathfrak{#1}}

\DeclareMathOperator*{\entails}{\vDash}
\DeclareMathOperator*{\nentails}{\nvDash}

\DeclareMathOperator*{\limply}{\rightarrow}

\newcommand{\stwop}{\Sigma_2^\tn{P}}

\newcommand{\papsd}{$P=(\fml{V},\fml{H},\fml{E},\fml{F})$\xspace}
\newcommand{\papdef}{$P=(\fml{V},\fml{H},\fml{E},\fml{F},c)$\xspace}

\definecolor{gray}{rgb}{.4,.4,.4}
\definecolor{midgrey}{rgb}{0.5,0.5,0.5}
\definecolor{middarkgrey}{rgb}{0.35,0.35,0.35}
\definecolor{darkgrey}{rgb}{0.3,0.3,0.3}
\definecolor{darkred}{rgb}{0.7,0.1,0.1}
\definecolor{midblue}{rgb}{0.2,0.2,0.7}
\definecolor{darkblue}{rgb}{0.1,0.1,0.5}
\definecolor{defseagreen}{cmyk}{0.69,0,0.50,0}
\definecolor{lightgreen}{HTML}{90EE90}

\newcommand{\jnote}[1]{\noindent$\llbracket$\textcolor{darkred}{joao}:~~{\sl\textcolor{darkgrey}{#1}}$\rrbracket$}

\newcommand{\anoteF}[1]{}
\newcommand{\jnoteF}[1]{}

\setlist{nolistsep}

\expandafter\def\expandafter\normalsize\expandafter{%
    \normalsize
    \setlength\abovedisplayskip{2.0pt}
    \setlength\belowdisplayskip{2.0pt}
    \setlength\abovedisplayshortskip{2.0pt}
    \setlength\belowdisplayshortskip{2.0pt}
}

\newcommand{\mailtodomain}[1]{\href{mailto:#1@ciencias.ulisboa.pt}{\textrm{\nolinkurl{#1}}}}

\newcommand{\frmeq}[1]{\begin{empheq}[box={\fboxsep=2.5pt\fbox}]{flalign*}#1\end{empheq}}



%% file: abs.tex
%
\begin{abstract}
  Recent years have witnessed a fast-growing interest in computing
  explanations for Machine Learning (ML) models predictions.
  %
  For non-interpretable ML models, the most commonly used approaches
  for computing explanations are heuristic in nature. In contrast,
  recent work proposed rigorous approaches for computing explanations,
  which hold for a given ML model and prediction over the entire
  instance space.
  This paper extends earlier work to the case of boosted trees
  and assesses the quality of explanations obtained with
  state-of-the-art heuristic approaches.
  On most of the datasets considered, and for the vast majority of
  instances, the explanations obtained with heuristic approaches
  are shown to be inadequate when the entire instance space is
  (implicitly) considered. 
  %
  %
\end{abstract}
%

%% file: intro.tex

\section{Introduction} \label{sec:intro}

\anoteF{1 page?}

Progress in Machine Learning (ML) has motivated efforts towards
verifying ML models properties
and
developing a better understanding of their outcomes.
As a result, two
concrete lines of research can be broadly identified.
One line is concerned with validating and ensuring specific properties
of neural networks. Another line is concerned with developing human
interpretable explanations for predictions made by ML models.
Perhaps unsurprisingly, both lines of research have witnessed a
growing use of logic-based
methods~\cite{barrett-cav17,kwiatkowska-cav17,kwiatkowska-ijcai18,darwiche-ijcai18,ipnms-ijcar18,inms-aaai19}.
%
The relevance of eXplainable Artificial Intelligence (XAI) is
illustrated by a fast growing number of works offering alternatives
into computing explanations for ML predictions.
More importantly, recent legislation imposes a requirement on the
explainability of ML systems~\cite{goodman-aimag17,eu-reg16}.

Some ML models are readily interpretable. This is the case with
logic-based models, e.g.\ decision trees, lists or
sets~\cite{leskovec-kdd16,rudin-icdm16,rudin-kdd17,nipms-ijcai18,ipnms-ijcar18}. Other
ML models are not readily interpretable. This is the case with Neural
Networks (NNs), Support Vector Machines (SVMs), and boosted trees,
among many others.
For models that are not readily interpretable, there has been work on
computing one or more explanations given an
instance~\cite{shavlik-nips95,kononenko-dke09,kononenko-jmlr10,muller-jmlr10,doshi-velez-nips15,gales-asru15,doshi-velez-ijcai17,muller-dsp18,rudin-aaai18,doshi-velez-aaai18a,doshi-velez-aaai18b,darwiche-ijcai18,kim-nips18a,jaakkola-nips18}.
%
%
One well-known approach for computing explanations is heuristic in
nature. Such explanations can be described as \emph{local}, i.e.\ the
computation of an explanation explores locally the instance sub-space
close to a given instance.
Well-known examples
are
LIME~\cite{guestrin-kdd16b} and, more recently,
Anchor~\cite{guestrin-aaai18}.
Since these approaches are local in nature, and so do not consider the
entire instance space, a natural question is to understand how reliable
the computed (local) explanations are.
For example, computed local explanations may be too optimistic, in
that there could exist instances (in instance space) for which the
computed explanation fails to apply, i.e.\ a different prediction is
obtained with the ML model. Alternatively, computed local explanations
may be too pessimistic, in that it may be possible to prove that some
literals in an explanation are irrelevant and can be dropped.
Recent work~\cite{guestrin-aaai18} compares Anchor against LIME, and
shows that the former is significantly more accurate than the latter.
However, and to our best knowledge, there is no earlier work assessing
the quality of the local explanations computed by Anchor or LIME
against some (global) reference.

Logic-based approaches have been proposed
recently\cite{darwiche-ijcai18,inms-aaai19}. 
They provide
strong guarantees given that computed explanations hold globally
over feature space, in contrast with local explanations computed with
heuristic approaches.
%
Shih et al.~\cite{darwiche-ijcai18} propose a compilation
based approach, representing all prime implicants of the function
explaining some prediction. Ignatiev et al.~\cite{inms-aaai19}
propose 
to compute prime implicants on demand, by formulating
the problem of computing an explanation as abductive reasoning.
Whereas the former approach enables aggregated analysis of
explanations, the latter approach is expected to scale better, as
explanations are computed on demand.

This paper builds on the approach of Ignatiev et
al.~\cite{inms-aaai19}, but investigates instead the computation of
global explanations for the concrete case of boosted trees. More
importantly, the paper develops solutions for assesssing the quality
of local explanations, using boosted trees as a test case. Overall,
the paper  has three main contributions.
First, the paper extends earlier work on finding global
explanations~\cite{inms-aaai19} to the case of boosted trees computed
with XGBoost~\cite{guestrin-kdd16a}\footnote{%
  XGBoost has achieved significant success in ML challenges hosted by
  \href{https://www.kaggle.com/}{Kaggle}.}, by devising a new
constraint-based encoding for boosted trees. As shown in the
experiments, computing restricted forms of abduction is far more
efficient on the proposed encoding of boosted trees than on the
original encoding of NNs~\cite{fischetti-cj18}.
Second, the paper develops algorithms for: (i) assessing the quality
of local explanations; (ii) repairing those local explanations when
they are optimistic; (iii) refining local explanations in case they
are pessimistic. The algorithms have been integrated in the XPlainer
XAI tool.
Third, 
the paper conducts the first experimental
assessment of the quality of explanations computed by Anchor and 
LIME in light of global explanations.
The paper considers five datasets
~\cite{feldman-kdd15,rudin-kdd17,guestrin-aaai18}, which are classified
with XGBoost~\cite{guestrin-kdd16a}. For two datasets, Anchor
is optimistic in more than 99\% of the instances,
meaning that the explanations computed by Anchor fail to apply for instances of
input space in more than 99\% of the cases.
%
For two other datasets, Anchor is optimistic in more than
80\% of the instances.
Although the results indicate that LIME is often more pessimistic than
Anchor, none of the tools dominates the other in terms of computing
optimistic explanations. These results offer more fine-grained
insights than earlier comparisons~\cite{guestrin-aaai18}.
%
%
%
Depending on the dataset considered, global explanations can be larger
than local explanations. This is a necessary result since 
global explanations are accurate (being either
subset- or cardinality-minimal) and so can neither be optimistic nor
pessimistic.
Furthermore, and for the boosted trees computed with
XGBoost~\cite{guestrin-kdd16a}, the run times of XPlainer are in
general comparable to those of LIME and Anchor.

The paper is organized as follows.
\autoref{sec:prelim} introduces the notation and definitions.
%
\autoref{sec:enc} develops an encoding for computing
global explanations with boosted tree classifiers.
\autoref{sec:cases} proposes algorithms behind the XPlainer tool.
These algorithms include finding subset- and cardinality-minimal
(global) explanations, validating heuristic explanation, repairing
heuristic explanations in case these are optimistic, and refining
heuristic explanations in case these are pessimistic.
\autoref{sec:res} analyzes experimental results obtained on five
well-known datasets~\cite{feldman-kdd15,rudin-kdd17,guestrin-aaai18}.
\autoref{sec:relw} offers a brief overview of related work and the
paper concludes in~\autoref{sec:conc}.

\jnoteF{ToDo:
  \begin{enumerate}
  \item Importance of XAI.
  \item What is an explanation?
    \begin{itemize}
    \item {\bf Note:} Must cover representative set of examples
    \end{itemize}
  \item Local vs. recently proposed global
    explanations~\cite{inms-aaai19}. 
  \item Claims:
    \begin{itemize}
    \item There exist models, of which boosted trees is the first
      example,  where a explanations based on a global view are in
      general close to computed local explanations.
    \end{itemize}
  \end{enumerate}
}


%% file: prelim.tex

\section{Background} \label{sec:prelim}

\anoteF{1--2.5 pages for preliminaries + existing approach}

%
A classification scenario is assumed, with categorical features
$\{f_1,\ldots,f_k\}$ and prediction classes $\{C_1,\ldots,C_m\}$. Each feature
$f_i$ takes values from some domain $\mathbb{D}_i$. (Features need not
be categorical, but this assumption simplifies the notation used.)
The training data consists of a set of \emph{instances}, where each
instance $I_j$ is taken from the \emph{instance space}, defined by
$\mathbb{D}_1\times\mathbb{D}_2\times\ldots\times\mathbb{D}_k$, and
where each instance $I_j$ is associated with some class $\pi_j$, taken
from $\{C_1,\ldots,C_m\}$, which is referred to as the target
prediction given the instance.

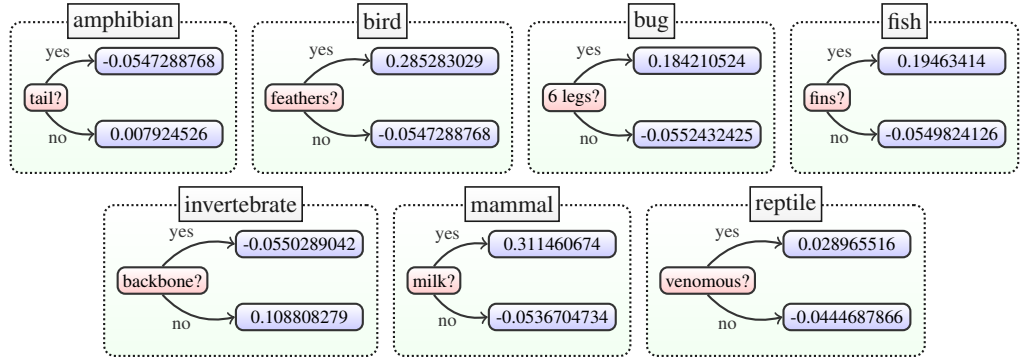
\begin{figure}[t]
  \begin{center}
    %
    \input{texfigs/ex1.tex}
    \caption{Example of a simplistic model trained by XGBoost. The
    model targets the well-known Zoo animal classification dataset.
    Here, the tree ensemble has 1 tree per each of the 7 classes with
    the depth of each tree being 1.}
    \label{fig:ex1}
  \end{center}
\end{figure}

\paragraph{Boosted Trees and Explanations.}
Boosted trees are one of the most widely used ML
models~\cite{guestrin-kdd16a}.
This paper considers XGBoost~\cite{guestrin-kdd16a}. Throughout the
paper, the well-known Zoo animal classification
dataset\footnote{\url{https://www.kaggle.com/uciml/zoo-animal-classification}}
is used as the running example. The result of running XGBoost on this
dataset is shown in~\autoref{fig:ex1}.
A larger number of tree nodes (or even more trees) could be considered
for each class. However, for the purposes of illustrating the main
ideas in the paper, the simpler version shown suffices.
\autoref{sec:enc} provides a more detailed account of boosted trees,
that will serve to motivate the development of constraint-based
encodings.

%
For the running example, 
one instance in the dataset is: 
\frmeq{
  \quad &
  \begin{array}{ll}
    \tn{\bf IF} &
    \eqlit{animal\_name}{pitviper}\land\nlit{hair}\land\nlit{feathers}\land\lit{eggs}\land\nlit{milk}\land\noarg
    \\
    &
    \nlit{airborne}\land\nlit{aquatic}\land\lit{predator}\land\nlit{toothed}\land\lit{backbone}\land\lit{breathes}\land\noarg
    \\
    &
    \lit{venomous}\land\nlit{fins}\land\eqlit{legs}{\tn{0}}\land\lit{tail}\land\nlit{domestic}\land\nlit{catsize}
    \\[2pt]
    \tn{\bf THEN} & \eqlit{class}{reptile}\\
  \end{array}
  &
}
Given the instance above, the execution of
Anchor~\cite{guestrin-aaai18} on the model shown produces the
following (local) explanation:
\frmeq{
  \quad &
  \begin{array}{ll}
    \tn{\bf IF} &
    \nlit{hair}\land\nlit{milk}\land\nlit{toothed}\land\nlit{fins} \\[2pt]
    \tn{\bf THEN} & \eqlit{class}{reptile} \\
  \end{array}
  &
}
Unfortunately, even for this simple dataset, and considering only the
instances in the original dataset, there is at least another instance
for which the Anchor explanation also applies, but which the boosted
tree predicts a \emph{different} class:
\frmeq{
  \quad &
  \begin{array}{ll}
    \tn{\bf IF} &
    \eqlit{animal\_name}{toad}\land\nhlit{hair}\land\nlit{feathers}\land\lit{eggs}\land\nhlit{milk}\land\noarg
    \\
    &
    \nlit{airborne}\land\nlit{aquatic}\land\nlit{predator}\land\nhlit{toothed}\land\noarg\lit{backbone}\land\lit{breathes}\land\noarg
    \\
    &
    \nlit{venomous}\land\nhlit{fins}\land\eqlit{legs}{4}\land\nlit{tail}\land\nlit{domestic}\land\nlit{catsize}
    \\[2pt]
    \tn{\bf THEN} & \eqlit{class}{\textcolor{red}{\bfseries{amphibian}}}\\
  \end{array}
  &
}
%
By analyzing the weight resulting from each
tree, we can conclude that the boosted tree prediction for this instance is indeed
\tl{amphibian}.
The remainder of this paper investigates approaches for assessing the
quality of the explanations computed by heuristic approaches, like LIME
and Anchor, but also for computing (global) explanations.

\paragraph{Logic-Related Concepts.}

Definitions standard in first-order logic (FOL) are
assumed~(e.g.~\cite{gallier-bk03}).
Given a signature $\fml{S}$ of predicate and function symbols, each of
which is characterized by its arity, a theory $\fml{T}$ is a set of
first-order sentences over $\fml{S}$.
$\fml{S}$ is extended with the predicate symbol $=$, denoting logical
equivalence\footnote{%
  Sorts could be used to add rigor to the presentation. However, to
  keep notation as simple as possible, sorts are omitted.}.
A model $\mathfrak{M}$ is a pair $\mathfrak{M}=(\fml{U},\fml{I})$, where
$\fml{U}$ denotes a universe, and $\fml{I}$ is an interpretation that
assigns a semantics to the predicate and function symbols of
$\fml{S}$.
A set $\fml{V}$ of variables is assumed, distinct from the symbols in
$\fml{S}$.
A (partial) assignment $\nu$ is a (partial) function from $\fml{V}$
to $\fml{U}$. Assignments are represented as conjunctions of
literals (or \emph{cubes}), where each literal is of the form $v=u$
s.t.\ $v\in\fml{V}$, $u\in\fml{U}$.
We use cubes and assignments interchangeably.
Whenever convenient, cubes are treated as sets of literals.
The set of free variables in a formula $\fml{F}$ is denoted by
$\tl{free}(\fml{F})$.
Assuming the standard semantics of FOL, and given an assignment $\nu$
and corresponding cube $C$, the notation
$\mathfrak{M},C\entails\fml{F}$ is used to denote that $\fml{F}$ is
true under model $\mathfrak{M}$ and cube $C$ (or assignment $\nu$). In
this case, $\nu$ (resp.~$C$) is called a satisfying assignment
(resp.~cube), and the assignment is partial if $|C|<|\fml{V}|$
(and so if $\nu$ is partial).
A solver for a FOL theory $\fml{T}$ is referred to as a
$\fml{T}$-oracle.

The generalization of prime implicants to FOL~\cite{marquis-fair91}
will be used throughout.
%
  Given a FOL formula $\fml{F}$ with a model
  $\mathfrak{M}=(\fml{U},\fml{I})$, a cube $C$ is a prime implicant of
  $\fml{F}$ if:
  (i) $\model{M},C\entails\fml{F}$; and
  (ii) if $C'$ is a cube with $\model{M},C'\entails\fml{F}$ and
    $\model{M},C'\entails C$, then $\model{M},C\entails C'$.
%
A smallest prime implicant is a prime implicant of minimum size.
Smallest prime implicants can be related with minimum satisfying
assignments~\cite{dillig-cav12}.
%
A prime implicant $C$ of $\fml{F}$ and $\model{M}$ given a cube
$C'$ is a prime implicant of $\fml{F}$ such that $C\subseteq C'$.

Satisfiability Modulo Theories (SMT) represent restricted (and often
decidable) fragments of FOL~\cite{barrett-hmc18,sebastiani-hdbk09}.
All the above definitions apply to SMT.
The ML models proposed in this paper exploit the decidable Linear Real
Arithmetic (LRA) fragment of FOL~\cite{barrett-hmc18}. The function
symbols are $\{+,-,\times\}$ and the predicate symbol is $\le$, with
the universe being $\mathbb{Q}$.
%
%

\paragraph{Abduction and Prime Implicants.}
Given some manifestations (e.g.\ a prediction), a set of hypotheses
(e.g.\ the given instance), and a background theory (e.g.\ the
encoding of some ML model), abduction is the problem of computing
subset-minimal or cardinality-minimal subsets of the hypotheses which
are consistent with the background theory and entail the
manifestation~\cite{eiter-jacm95,jarvisalo-kr16,imms-ecai16}.
The relationship of abduction with prime implicants in the context of
computing explanations of ML models was established in earlier
work~\cite{inms-aaai19}. As a result, this paper considers solely
prime implicants as the desired explanations of predictions of ML
models.
%
%
As in earlier work~\cite{inms-aaai19}, we associate a logic theory
$\fml{T}$ with a given ML model $\mathbb{M}$, and encode $\mathbb{M}$
as a formula $\fml{M}$ of $\fml{T}$. Thus, in contrast with other
approaches~\cite{guestrin-kdd16b,guestrin-aaai18}, we must be able to
have access to a constraint-based representation (i.e.\ a formula)
$\fml{M}$ of the ML model $\mathbb{M}$.

We consider an instance $I$ with which a prediction $\pi$ is
associated. With a slight abuse of notation, $I$ is also used to
denote the cube associated with the instance, and $\pi$ is used to
denote the literals associated with prediction.
The relationship between abductive explanations and prime implicants
is well-known (e.g.~\cite{marquis-fair91,marquis-hdrums00}.
Regarding the computation of abductive explanations,
$I\land\fml{M}\nentails\bot$ and the same holds for any
subset of $I$. This means that it suffices to consider the constraint
$I\land\fml{M}\entails\pi$, which is equivalent to
$I\entails(\fml{M}\to\pi)$. Thus, a \emph{subset-minimal
  explanation}
$I_m$ (given $I$) is a prime implicant of $\fml{M}\to\pi$ (given
$I$), and a \emph{cardinality-minimal explanation} $I_M$ (given $I$)
is a cardinality-minimal prime implicant of $\fml{M}\to\pi$ (given
$I$).
Hence, we can compute subset-minimal (resp.~cardinality-minimal)
explanations by computing instead prime implicants (resp.~shortest
PIs) of $\fml{M}\to\pi$.
As a final remark, the cardinality minimal prime implicants of
$\fml{M}\to\pi$ are selected among those that are contained in
$I$.
For instance, assuming a FOL encoding of a boosted tree (this encoding
is detailed in~\autoref{sec:enc}), and given the Zoo running example,
and the instance yielding the \tl{reptile} prediction, then we can
compute the explanation (as described in~\autoref{sec:cases}):
\frmeq{
  \quad &
  \begin{array}{ll}
    \tn{\bf IF} &
    \nlit{feathers}\land\nlit{milk}\land\lit{backbone}\land\nlit{fins}\land\eqlit{legs}{\tn{0}}\land\lit{tail}
    \\[2pt]
    \tn{\bf THEN} & \eqlit{class}{reptile} \\
  \end{array}
  &
}
We emphasize that this explanation is a prime implicant of
$\fml{M}\limply\pi$. Thus, and by definition, the explanation
\emph{guarantees} that the prediction remains unchanged for \emph{any}
other instance in instance space for which the six literals 
remain unchanged. A  downside is that this explanation can include
more literals that the ones computed by Anchor~\cite{guestrin-aaai18}.


\jnoteF{Include modified pseudo-code from AAAI'19 paper.}

\paragraph{Computing Abductive Explanations.}
\input{abd-alg}
Given the formalization abo\-ve, abductive explanations can be
obtained by computing prime implicants.
Earlier work~\cite{inms-aaai19} outlined two algorithms for computing
explanations, based on the extraction of prime implicants and
(smallest) prime implicants. The former corresponds to subset-minimal
explanations and is shown in~\autoref{alg:smexpl}. In contrast, the
latter (see~\cite[Algorithm~2]{inms-aaai19}) corresponds to
cardinality-minimal explanations.
From a computational complexity viewpoint, and assuming as oracle for
NP either an ILP or LRA solver, computing subset-minimal explanations
is hard for NP, and can be solved with a linear number of calls to an
oracle for NP~\cite{umans-tcad06} (as shown in~\autoref{alg:smexpl}).
In contrast, computing a cardinality-minimal explanation is (believed
to be) harder, being hard for $\stwop$, and can be solved with a
linear number of calls to an oracle for $\stwop$~\cite{umans-tcad06}
(or alternatively, using implicit hitting sets as shown
in~\cite[Algorithm~2]{inms-aaai19}). 

%% file: texfigs/ex1.tex
%
%
%

\tikzstyle{box} = [draw=black!90, thick, rectangle, rounded corners,
                     inner sep=10pt, inner ysep=20pt, densely dotted,
                     top color=white,bottom color=green!5
                  ]
\tikzstyle{title} = [draw=black!90, fill=black!5, semithick, top color=white,
                     bottom color = black!5, text=black!90, rectangle,
                     font=\small, inner sep=2pt, minimum height=1.3em
                    ]
\tikzstyle{feature} = [rectangle,font=\scriptsize,rounded corners=1mm,thick,%
                       draw=black!80, top color=white,bottom color=red!20,%
                       draw, minimum height=1.0em, text centered,%
                       inner sep=2pt%
                      ]
\tikzstyle{score} = [rectangle,font=\scriptsize,rounded corners=1mm,thick,%
                     draw=black!80, top color=white,bottom color=blue!20,%
                     draw,text width=4.4em, minimum height=1.0em, text centered,%
                     inner sep=2pt%
                    ]

\begin{adjustbox}{center}
\begin{tabular}{cccc}
    \begin{tikzpicture}[node distance = 4.0em, auto]
        \node [box] (box) {%
        \begin{minipage}[t!]{0.188\textwidth}
            \vspace{0.6cm}\hspace{1.5cm}
        \end{minipage}
        };
        \node[title] at (box.north) {amphibian};

        \node [feature] (feat) at (-1.03, 0) {tail?};

        \node [score, above right = 0.3em and 1.0em of feat] (pos) {-0.0547288768};
        \node [score, below right = 0.3em and 1.0em of feat] (neg) { 0.007924526};

        \draw [->,thick,black!80] (feat.north) to[bend left ] node[above, pos=0.3, font=\scriptsize] {yes} (pos.west);
        \draw [->,thick,black!80] (feat.south) to[bend right] node[below, pos=0.3, font=\scriptsize] { no} (neg.west);
    \end{tikzpicture}
    &
    \begin{tikzpicture}[node distance = 4.0em, auto]
        \node [box] (box) {%
        \begin{minipage}[t!]{0.225\textwidth}
            \vspace{0.6cm}\hspace{1.5cm}
        \end{minipage}
        };
        \node[title] at (box.north) {bird};

        \node [feature] (feat) at (-1.04, 0) {feathers?};

        \node [score, above right = 0.3em and 1.0em of feat] (pos) { 0.285283029};
        \node [score, below right = 0.3em and 1.0em of feat] (neg) {-0.0547288768};

        \draw [->,thick,black!80] (feat.north) to[bend left ] node[above, pos=0.3, font=\scriptsize] {yes} (pos.west);
        \draw [->,thick,black!80] (feat.south) to[bend right] node[below, pos=0.3, font=\scriptsize] { no} (neg.west);
    \end{tikzpicture}
    &
    \begin{tikzpicture}[node distance = 4.0em, auto]
        \node [box] (box) {%
        \begin{minipage}[t!]{0.21\textwidth}
            \vspace{0.6cm}\hspace{1.5cm}
        \end{minipage}
        };
        \node[title] at (box.north) {bug};

        \node [feature] (feat) at (-1.03, 0) {6 legs?};

        \node [score, above right = 0.3em and 1.0em of feat] (pos) { 0.184210524};
        \node [score, below right = 0.3em and 1.0em of feat] (neg) {-0.0552432425};

        \draw [->,thick,black!80] (feat.north) to[bend left ] node[above, pos=0.3, font=\scriptsize] {yes} (pos.west);
        \draw [->,thick,black!80] (feat.south) to[bend right] node[below, pos=0.3, font=\scriptsize] { no} (neg.west);
    \end{tikzpicture}
    &
    \begin{tikzpicture}[node distance = 4.0em, auto]
        \node [box] (box) {%
        \begin{minipage}[t!]{0.19\textwidth}
            \vspace{0.6cm}\hspace{1.5cm}
        \end{minipage}
        };
        \node[title] at (box.north) {fish};

        \node [feature] (feat) at (-1.03, 0) {fins?};

        \node [score, above right = 0.3em and 1.0em of feat] (pos) {0.19463414};
        \node [score, below right = 0.3em and 1.0em of feat] (neg) {-0.0549824126};

        \draw [->,thick,black!80] (feat.north) to[bend left ] node[above, pos=0.3, font=\scriptsize] {yes} (pos.west);
        \draw [->,thick,black!80] (feat.south) to[bend right] node[below, pos=0.3, font=\scriptsize] { no} (neg.west);
    \end{tikzpicture}
\end{tabular}
\end{adjustbox}

\begin{adjustbox}{center}
\begin{tabular}{ccc}
    \begin{tikzpicture}[node distance = 4.0em, auto]
        \node [box] (box) {%
        \begin{minipage}[t!]{0.24\textwidth}
            \vspace{0.6cm}\hspace{1.5cm}
        \end{minipage}
        };
        \node[title] at (box.north) {invertebrate};

        \node [feature] (feat) at (-1.04, 0) {backbone?};

        \node [score, above right = 0.3em and 1.0em of feat] (pos) {-0.0550289042};
        \node [score, below right = 0.3em and 1.0em of feat] (neg) {0.108808279};

        \draw [->,thick,black!80] (feat.north) to[bend left ] node[above, pos=0.3, font=\scriptsize] {yes} (pos.west);
        \draw [->,thick,black!80] (feat.south) to[bend right] node[below, pos=0.3, font=\scriptsize] { no} (neg.west);
    \end{tikzpicture}
    &
    \begin{tikzpicture}[node distance = 4.0em, auto]
        \node [box] (box) {%
        \begin{minipage}[t!]{0.2\textwidth}
            \vspace{0.6cm}\hspace{1.5cm}
        \end{minipage}
        };
        \node[title] at (box.north) {mammal};

        \node [feature] (feat) at (-1.04, 0) {milk?};

        \node [score, above right = 0.3em and 1.0em of feat] (pos) {0.311460674};
        \node [score, below right = 0.3em and 1.0em of feat] (neg) {-0.0536704734};

        \draw [->,thick,black!80] (feat.north) to[bend left ] node[above, pos=0.3, font=\scriptsize] {yes} (pos.west);
        \draw [->,thick,black!80] (feat.south) to[bend right] node[below, pos=0.3, font=\scriptsize] { no} (neg.west);
    \end{tikzpicture}
    &
    \begin{tikzpicture}[node distance = 4.0em, auto]
        \node [box] (box) {%
        \begin{minipage}[t!]{0.245\textwidth}
            \vspace{0.6cm}\hspace{1.5cm}
        \end{minipage}
        };
        \node[title] at (box.north) {reptile};

        \node [feature] (feat) at (-1.04, 0) {venomous?};

        \node [score, above right = 0.3em and 1.0em of feat] (pos) {0.028965516};
        \node [score, below right = 0.3em and 1.0em of feat] (neg) {-0.0444687866};

        \draw [->,thick,black!80] (feat.north) to[bend left ] node[above, pos=0.3, font=\scriptsize] {yes} (pos.west);
        \draw [->,thick,black!80] (feat.south) to[bend right] node[below, pos=0.3, font=\scriptsize] { no} (neg.west);
    \end{tikzpicture}
\end{tabular}
\end{adjustbox}


%% file: abd-alg.tex

\SetKwBlock{Begin}{\texttt{begin}}{\texttt{end}}
\SetKw{Return}{\texttt{return}}
\SetKwIF{If}{ElseIf}{Else}{\texttt{if}}{\texttt{:}}{\texttt{elif}}{\texttt{else:}}{}%
\SetKwFor{ForEach}{\texttt{for}}{\texttt{:}}{}%
\SetKwFor{While}{\texttt{while}}{\texttt{:}}{}%

\SetKw{KwNot}{not\xspace}
\SetKw{KwAnd}{and\xspace}
\SetKw{KwOr}{or\xspace}
\SetKw{KwBreak}{break\xspace}
\SetKwData{false}{False}
\SetKwData{true}{True}
\SetKwData{st}{{\slshape st}}
\SetKwData{cores}{$\mathcal{C}$}
\SetKwFunction{ent}{Entails}
\SetKwFunction{ip}{IP}
\SetKwBlock{Let}{let}{end}
\SetKwBlock{FBlock}{}{end}

\SetKwFunction{SAT}{SAT}
\SetKwFunction{CNF}{CNF}
\SetKwFunction{minhs}{MinimumHS}
\SetKwFunction{model}{GetAssignment}
\SetKwFunction{falselits}{PickFalseLits}

\begin{algorithm}[!t]
  \caption{Computing a subset-minimal explanation} \label{alg:smexpl}

  \DontPrintSemicolon
  \SetAlgoNoLine
  \LinesNumbered
  \SetFillComment

  \KwIn{\,\,\,\,formula $\fml{M}$, input $I$, prediction $\pi$}
  \KwOut{Subset-minimal explanation $I_m\subseteq I$}
  \BlankLine
  $I_m\gets I$\;
  \ForEach{$f \in I_m$}{
    \If{$\ent(I_m\setminus\{f\},\fml{M}\rightarrow\pi)$}{
      $I_m\gets I_m\setminus\{f\}$
    }
  }
  \Return $I_m$ \;
  \BlankLine
\end{algorithm}

%% file: enc.tex
\section{Encoding Boosted Trees} \label{sec:enc}



This section proposes an SMT encoding of an ensemble of decision trees
produced by XGBoost algorithm. 
%
Suppose our training data is specified over $k$ features, $f_1,\ldots,f_k$, and there are $m$ possible classification outcomes.
For example, there are 17 features per sample in the Zoo dataset. Features describe characteristics of an animal, e.g. whether an animal lays eggs, the number of legs, etc. There are seven possible outcomes: amphibian,  bird, bug, invertebrate, fish, mammal, and reptile (see Figure~\ref{fig:ex1}).
For simplicity, we assume that all features are binary. We discuss how to extend our encoding  to categorical and continuous features in the end of the section.

An XGBoost model is an ensemble of decision trees.  A decision tree is a binary tree. A node of a tree is denoted by $n_i$. We distinguish between non-leaf or internal nodes and leaf nodes.   A non-leaf node $n_i$ of a decision tree contains a logical predicate over a feature variable of the form $(f_{j} \text{ is true?})$ or $(f_{j}?)$ for short~\footnote{XGBoost uses constraints of the form $(f_{j} < 0.5)$ which is equivalent to $(\neg{f_j}?)$ for binary features.}. Outgoing edges of a node are labeled \emph{true} (the right branch) and \emph{false} (the left branch).
For convenience, we assume that an internal node has two attributes $pred$ and $idx$:
$n_i.pred$ stores the predicate of this node and $n_i.idx$ stores the index of the feature variable in $n_i.pred$.  A leaf node $n_i$ contains a numerical value $v$, $v \in \mathbb{R}$. We assume that a leaf node has one attribute $n_i.val$ that stores this value $v$.  Consider the first tree in the Zoo example. The tree has  three nodes.  The root node $n_0$ has the following attributes: $n_0.pred = (tail?)$ and $n_0.idx = tail$.
The first leaf node $n_1$ has one attribute $n_1.val = -0.0547$ and the second
leaf node $n_2$ has one attribute $n_2.val = 0.0079$.

The number of trees in the ensemble $T$ is equal to the number of classes, $m$, times the number of trees per class, $q$, where $q$ is specified by the user. In other words, we have $mq$ trees, $T = \{t_1,\ldots, t_{mq}\}$, where the $j$th class is represented by $q$ trees, $\{t_{qj+1},\ldots, t_{q(j+1)}\}$. The depth of a tree can also be specified by the user. In our Zoo example, $q$ is one and the depth of the tree is one.

Next, we consider how classification is performed using the XGBoost model. Given a concrete example $s$, we want to know which class it belongs to.  To answer this question, we compute the score of the $j$th class for $s$, $j \in [m]$. W.l.o.g. we discuss how to find the score of the first class. We consider $q$ trees that represent the first class in the ensemble $t_{l}$, $l\in[q]$. Each of these trees contributes a score to the total score of the first class.  We consider each tree individually. A \emph{prediction path} of $s$ in $t_{l}$ is a path $p = (n_0^{l}, \ldots, n_d^{l})$ from the root $n_0^{l}$ to a leaf $n_d^{l}$ such that for an internal node $n_i$ on the path $n_i.pred$ holds if $p$ follows the right branch and does not hold if $p$ follows the left branch. The leaf $n_d^{l}$ contains the score value. Then we aggregate the result from $q$ trees, $v_1 = \sum_{l=1}^q n_d^{l}.val$ to get the final score of the $1$st class. We perform the same score computation for all classes. The class with the largest score wins. In general, we can normalize these scores to obtain probabilities but scores are sufficient for our purpose. Consider our running example with the first instance from \autoref{sec:prelim} classifed as reptile.
We have seven trees here
, as $q=1$.
So, we get $v_1 = -0.0547$ as $\lit{tail}$ holds,  $v_2 = -0.0547$ as $\lit{feathers}$ does not hold.
Similarly, we get $v_3 = -0.0552$, $v_4 = -0.0549$,
$v_5 = -0.0550$, $v_6 = -0.0537$ and $v_7 = 0.0290$. As can be seen, $v_7$ has the largest score so the classification class is ``reptile''.

Next, we consider how to encode an XGBoost model into SMT. At a high level, our encoding simulates the scores computation for a possible input.  We introduce three sets of variables. For a binary feature $f_i$ we introduce a Boolean variable $b_i$, $i \in [k]$. These Boolean variables represent the space of all possible inputs. For the $l$th tree we introduce a real valued variable $r_l$, $l \in [mq]$. The $r_l$ variable encodes the score contribution from the $l$th tree. Finally, for the $j$th class we introduce one variable $v_j$, $j\in[m]$ that stores the score of the $j$th class.
We connect Boolean variables with predicates: $b_i\leftrightarrow (f_i?)$. Then, we encode the score computation for a tree $t_l \in T$ by
encoding all paths in $t_l$ as follows. Let $P(t_l)$ be a set of all distinct paths from the root to a leaf in $t_l$. Consider a path $p \in P(t_l)$, $ p = (n_0, \ldots, n_d)$ from the root $n_0$ to a leaf $n_d$. We recall that if $p$ follows the right (left, resp.) branch in a node $n_i$ then the predicate $n_i.pred$ has to hold (to be violated, resp.).  Let $R_p$ be a set of nodes where $p$ takes the right branch and $L_p$ be a set of nodes where $p$ takes the left branch.  We enforce the following constraints:
\begin{align*}
\bigwedge_{n_i \in R_p} b_{n_i.idx}  \bigwedge_{n_i \in L_p} \neg b_{n_i.idx} \rightarrow r_l = n_d.val,  \ \  p \in P(t_l), t_l \in T.
\end{align*}
%
To compute the score of the $j$th class we add $m$ constraints,  $j \in [m]$:
$v_j = \sum_{l=1}^q r_{qj+l}$.

Consider how the encoding works on the running example.
Consider the first tree.
For simplicity, we denote the index attribute of the root node as `tail'.
 We have two paths in the tree.
For the first path, we add a constraint
$b_{\text{tail}} \rightarrow r_1 = -0.0547$
and for the second path we add a constraint
$\nlit{b}_{\text{tail}} \rightarrow r_1 = 0.0079$.
With one tree per class, we get $v_1= r_1$. Other trees could be
encoded similarly.
%

Next we discuss how to extend our encoding to categorical and continuous data. In case of categorical data, a common approach is to apply \emph{one hot encoding} to convert discrete values to binary values. This transformation is performed on the original data. Hence, our encoding can be applied directly with a small augmentation. We enforce that exactly one of binary features that encode a categorical feature can be true.
The case of continuous features is handled similarly. The main difference is that logical predicates in a node are of the form $f_i < c$, where $c$ is a constant value. Here, for each predicate that occurs in $t_l, t_l \in T$, we introduce $b_{ic}$ Boolean variable such that
$b_{ic} \leftrightarrow (f_i < c)$. Then the encoding above can be reused.

%
%


%% file: cases.tex

\section{Reasoning about Explanations} \label{sec:cases}

This section discusses the practicality of the abduction-based
approach~\cite{inms-aaai19} and focuses on applying it to explanation
of a tree ensemble model using the novel constraints-based encoding
proposed above\footnote{The ideas of this section will still apply if
  another encoding of a tree ensemble is considered.}.
Hereinafter, this approach is referred to as \emph{XPlainer}.
Concretely, the section outlines possible use cases of applying
Xplainer in practice: either alone or together with a heuristic
explanation approach, e.g.\ LIME or Anchor.

\input{abd}

\input{valid}
\input{fix}
\input{reduc}


%% file: abd.tex

\subsection{Minimal Global Explanations} \label{sec:abd}

First of all, the XPlainer approach can be applied directly to
computing subset- and cardinality-minimal
explanations~\cite{inms-aaai19} for boosted trees using the encoding
proposed in \autoref{sec:enc}.
Let us apply XPlainer to the running example model shown in
\autoref{fig:ex1}.
Assume that the encoding of the model is represented as a formula
$\fml{M}$, which is the following conjunction of constraints.
%
%
\begin{equation*}
  \arraycolsep=1.4pt
  \fml{M}=\left\{
    \begin{array}{llrlllrlc}
      (b_{\text{tail}} & \rightarrow r_1= & -0.0547) & \land &
      (\neg{b_{\text{tail}}} & \rightarrow r_1= & 0.0079) & \land &
      (v_1=r_1) \\
      (b_{\text{feath}} & \rightarrow r_2= & 0.2853) & \land &
      (\neg{b_{\text{feath}}} & \rightarrow r_2= & -0.0547) & \land &
      (v_2=r_2) \\
      (b_{\text{6legs}} & \rightarrow r_3= & 0.1842) & \land &
      (\neg{b_{\text{6legs}}} & \rightarrow r_3= & -0.0552) & \land &
      (v_3=r_3) \\
      (b_{\text{fins}} & \rightarrow r_4= & 0.1946) & \land &
      (\neg{b_{\text{fins}}} & \rightarrow r_4= & -0.0549) & \land &
      (v_4=r_4) \\
      (b_{\text{bbone}} & \rightarrow r_5= & -0.0550) & \land &
      (\neg{b_{\text{bbone}}} & \rightarrow r_5= & 0.1088) & \land &
      (v_5=r_5) \\
      (b_{\text{milk}} & \rightarrow r_6= & 0.3615) & \land &
      (\neg{b_{\text{milk}}} & \rightarrow r_6= & -0.0537) & \land &
      (v_6=r_6) \\
      (b_{\text{venom}} & \rightarrow r_7= & 0.0290) & \land &
      (\neg{b_{\text{venom}}} & \rightarrow r_7= & -0.0445) & \land &
      (v_7=r_7) \\
    \end{array}
  \right\}
\end{equation*}
Let $l_i$ be a literal over Boolean variable $b_i$ used above, e.g.\
$l_i$ is either $b_i$ or $\neg{b_i}$.
The ``translation'' of each feature value $f_i\in I$ into the
corresponding literal $l_i$ is straightforward\footnote{In practice,
  categorical feature $\lit{legs}$ should be one-hot encoded. But for
  the sake of simplicity and without loss of correction, we use a
Boolean variable $b_{\text{6legs}}$, s.t.\ $b_{\text{6legs}}$ is true
iff $\lit{legs}=6$.}.
Now, for each input $I$ defined as a conjunction of literals over
$b_i$, the prediction is determined by the largest score value $v_j$,
$j\in[7]$, computed using formula $\fml{M}$.
Given the list of class scores $v_j$, the prediction of class $j$ can
be guaranteed using a conjunction of linear inequalities enforcing
value $v_j$ to be the largest, i.e.\ with the use of formula
$\pi_j=\bigwedge_{i\neq j}{v_j>v_i}$.
Consider the following input instance and its respective prediction
\frmeq{
  \quad &
  \begin{array}{ll} \tn{\bf IF} &
    \eqlit{animal\_name}{bear}\land\lit{hair}\land\nlit{feathers}\land\nlit{eggs}\land\lit{milk}\land\nlit{airborne}\land\noarg
    \\ &
    \nlit{aquatic}\land\lit{predator}\land\lit{toothed}\land\lit{backbone}\land\lit{breathes}\land\lit{venomous}\land\noarg
    \\
    &
    \nlit{fins}\land\eqlit{legs}{4}\land\nlit{tail}\land\nlit{domestic}\land\nlit{catsize}
  \\[2pt] \tn{\bf THEN} & \eqlit{class}{\bfseries{mammal}}\\
  \end{array} &
}
Since $\lit{mammal}$ represents class 6, this prediction can be encoded
as the following conjunction of inequalities (observe that they hold
for the considered input):
\begin{equation*}
  \pi_6 = \bigwedge_{i\in[7], i\neq 6}{v_6 > v_i}
\end{equation*}

Let us illustrate the flow of \autoref{alg:smexpl}.
The algorithm makes calls to a reasoner deciding whether or not a
candidate subset $I'$ of the input instance $I$ is a prime implicant
of $\fml{M}\rightarrow\pi$.
This is true iff formula $I'\land\fml{M}\land\neg{\pi}$ is
\emph{unsatisfiable}.
Hence, if we fix the values of features in $I'$ then \emph{no
misclassification}, i.e.\ $\neg{\pi}$, is \emph{possible} for model
$\fml{M}$.

Clearly, when $I_m$ includes all literals $l_i\in I$, formula
$I_m\land\fml{M}\land\neg{\pi}$ is unsatisfiable.
Recall that \autoref{alg:smexpl} iteratively removes literals $l_i$
from $I_m$ and checks whether or not
$I_m\setminus\{l_i\}\land\fml{M}\land\neg{\pi}$ is still
unsatisfiable.
If it is, $I_m\setminus\{l_i\}$ is still an implicant of
$\fml{M}\rightarrow\pi$, i.e.\ literal $l_i$ is not responsible for
the prediction $\pi$.
Otherwise, $I_m$ is not an implicant, i.e.\ $l_i$ is necessary and,
thus, must be included in the explanation.
This process repeatedly checks all literals $l_i\in I$.

One straightforward optimization to make before executing
\autoref{alg:smexpl} is to discard from $I_m$ all features unused by
the model as they cannot affect the prediction.
In our example, we can safely remove all features except for the seven
features used in the model.
This results in $I_m=\{\neg{b_{\text{feath}}}, b_{\text{milk}},
b_{\text{bbone}}, \neg{b_{\text{venom}}},
\neg{b_{\text{fins}}}, \neg{b_{\text{6legs}}},
\neg{b_{\text{tail}}}\}$.
Hence, the first literal to be tested by \autoref{alg:smexpl} is
$\neg{b_{\text{feath}}}$.
The reasoning oracle is called to check unsatisfiability of
$I_M\setminus\{\neg{b_{\text{feath}}}\}\land\fml{M}\land\neg{\pi}$.
Note that this formula is indeed unsatisfiable because the largest
class score is still $v_6=0.3615$, which is enforced by literal
$b_{\text{milk}}$.
Thus, $\neg{b_{\text{feath}}}$ is not crucial for the prediction and
it gets removed from $I_m$.
The second literal to check is $b_{\text{milk}}$.
This time, the oracle tests unsatisfiability of
$I_m\setminus\{\neg{b_{\text{feath}}},b_{\text{milk}}\} \land \fml{M}
\land \neg{\pi}$ and returns true.
This means that a misclassification can occur if $b_{\text{milk}}$ is
discarded.
Indeed, since variables $b_{\text{feath}}$ and $b_{\text{milk}}$ are
free, the oracle can assign any values to them, e.g.\ setting
$b_{\text{feath}}$ and $\neg{b_{\text{milk}}}$ results in $v_2=0.2853$
being the largest class score.
As a result, literal $b_{\text{milk}}$ is vital for the prediction to
persist.
The algorithm proceeds doing similar checks with respect to all the
remaining literals in $I_m$.
As a result, it ends up having $I_m=\{b_{\text{milk}}\}$, i.e.\ the
explanation contains only one feature.

Observe that explanations computed this way are subset-minimal.
Furthermore, since a reasoner deals with the properties of the
classifier's symbolic representation $\fml{M}$ in the \emph{complete
instance space}, these explanations are \emph{global}, i.e.\ they hold
for the entire space.
Given a global explanation $I_m$ for the prediction $\pi$ of a data
input $I$, it is guaranteed that there is no point $I'$ in the
instance space s.t.\ (1)~$I_m\subseteq I'$ and (2)~the prediction for
$I'$ is $\pi'\neq\pi$.
Global explanations are significantly more powerful than explanations
offered by the state-of-the-art heuristic approaches, e.g.
LIME~\cite{guestrin-kdd16a} or Anchor~\cite{guestrin-aaai18}, since
the latter ones hold only for a \emph{local neighborhood} of a given
instance.

As detailed in~\cite{inms-aaai19}, cardinality-minimal explanations
can also be computed, e.g.\ using the implicit hitting
approach~\cite{karp-soda11,imms-ecai16}.
Similarly to the case of subset-minimal explanations, one would need
to make a number of similar unsatisfiability calls to a reasoner.
However and in contrast to \autoref{alg:smexpl}, computing a smallest
size explanation is hard for $\stwop$ and in the worst case requires
an exponential number of iterations.

%% file: valid.tex
\subsection{Validating Heuristic Explanations} \label{sec:valid}

Besides computing global explanations directly, XPlainer can be
applied to validating given heuristic explanations.
Indeed, one can immediately notice that in order to check the validity
of a heuristic explanation $I_h$ for a model formula $\fml{M}$ and
data instance $I$ classified as $\pi$, it suffices to do one oracle
call similar to the ideas outlined in \autoref{sec:abd}.
The corresponding procedure is shown in \autoref{alg:valid}.

\SetKwBlock{Begin}{\texttt{begin}}{\texttt{end}}
\SetKw{Return}{\texttt{return}}
\SetKwIF{If}{ElseIf}{Else}{\texttt{if}}{\texttt{:}}{\texttt{elif}}{\texttt{else:}}{}%
\SetKwFor{ForEach}{\texttt{for}}{\texttt{:}}{}%
\SetKwFor{While}{\texttt{while}}{\texttt{:}}{}%

\SetKw{KwNot}{\texttt{not}\xspace}
\SetKw{KwAnd}{and\xspace}
\SetKw{KwOr}{or\xspace}
\SetKw{KwBreak}{break\xspace}
\SetKwData{false}{False}
\SetKwData{true}{True}
\SetKwData{st}{{\slshape st}}
\SetKwData{cores}{$\mathcal{C}$}
\SetKwFunction{ent}{Entails}
\SetKwFunction{ip}{IP}
\SetKwBlock{Let}{let}{end}
\SetKwBlock{FBlock}{}{end}

\SetKwFunction{SAT}{SAT}
\SetKwFunction{CNF}{CNF}
\SetKwFunction{minhs}{MinimumHS}
\SetKwFunction{model}{GetAssignment}
\SetKwFunction{falselits}{PickFalseLits}
\SetKwFunction{extractvals}{ExtractValues}

\begin{algorithm}[!t]
  \caption{Validating a heuristic explanation} \label{alg:valid}

  \DontPrintSemicolon
  \SetAlgoNoLine
  \LinesNumbered
  \SetFillComment

  \KwIn{\,\,\,\,formula $\fml{M}$, input $I$, prediction $\pi$, and explanation $I_h$}
  \KwOut{counterexample $C$ for explanation $I_h$}
  \BlankLine
  $C \gets \emptyset$\;
  \If{\KwNot $\ent(I_h,\fml{M}\rightarrow\pi)$}{
    $\mu\gets\model()$\;
    $C\gets\extractvals(\mu)$
  }
  \Return $C$ \;
  \BlankLine
\end{algorithm}

As later shown in \autoref{sec:res}, this simple and efficient
procedure is able to prove or disprove an explanation to be globally
correct.
(An example of an explanation reported by Anchor and a counterexample
computed by \autoref{alg:valid} is discussed in \autoref{sec:prelim}.)
Since heuristic approaches compute local explanations, it is not
surprising that most of them are incorrect from the perspective of the
complete instance space (see \autoref{sec:res}).
An upside of XPlainer is that it can efficiently provide a
counterexample to an explanation demonstrating its unsoundness.
Moreover, it can be used not only to (in)validate an explanation by
providing one counterexample, but it can also enumerate (all or a
limited number of) counterexamples showing \emph{why} and \emph{when}
the explanation is incorrect.
Based on such evidence, one can try to devise a way to correct the
explanation or compute a better alternative from scratch.

In many settings, computing correct explanations is crucial from a
practitioner's point of view as they are supposed to provide a user
with hints of why the model behaves one way or another.
These hints should reflect the real properties of the model.
If they do not, a comprehensive understanding of the model is
infeasible.


%% file: fix.tex

\subsection{Repairing Heuristic Approaches} \label{sec:fix}

If an explanation is proved to be too optimistic, it is often
vital to find a way to make a number of (ideally, \emph{minimal})
changes to the explanation so that it becomes correct in the
instance space.
An explanation $I_h$ for the prediction of instance $I$ is optimistic
when the features of $I_h$ \emph{do not suffice} to guarantee the
prediction.
A way to repair $I_h$ is to find another subset of features
$I_m\subseteq I$ such that $I_m$ is a correct explanation.
It is preferred to minimize the ``distance'' between $I_h$ and $I_m$.
This is another task where XPlainer can help since it deals with a
logical representation of the classifier and is able to answer queries
about the classifier system and its behavior.

Computing minimum size changes to the explanation is related to
identifying minimal inconsistencies and/or diagnoses for a failing
system subject to user preferences, which has been studied in prior
works~\cite{boutilier-jair04,giunchiglia-aicom13,msp-sat14}.
It is known, however, that in a number of settings the latter problem
is hard for the second level of the polynomial
hierarchy~\cite{msp-sat14}.
Therefore, it seems unlikely that given an explanation, one can
efficiently extract another one, which would be guaranteed to
minimally differ from the original one.
However, the problem can be solved heuristically.
An approach to this problem using the abilities of XPlainer is shown
in \autoref{alg:repair}.
The algorithm follows the procedure of \autoref{alg:smexpl} for
extracting subset-minimal explanations.
It additionally receives an (invalid) heuristic explanation $I_h$ that
is to be repaired.
The key idea of the approach is to \emph{delay} as much as possible
the testing of features of $I_h$ while computing a valid explanation.
Hence, the algorithm tries to remove as many features from the outside
of $I_h$ as possible.
Afterwards, it traverses the features of $I_h$.
To emphasize again, \autoref{alg:repair} does not guarantee the result
explanation to minimally differ from the original explanation.
However, an upside of the algorithm is that it does not deal with a
$\stwop$ problem --- instead, it makes a linear number of calls to an
NP-oracle, which is practically much more efficient.
Having such a repair should suffice in many practical situations.

As an example, recall the $\lit{pitviper}$ instance and the invalid
explanation of Anchor shown in \autoref{sec:prelim}.
Anchor claims features $\nlit{hair}$, $\nlit{milk}$, $\lit{toothed}$,
and $\nlit{fins}$ to be responsible for the $\lit{reptile}$ prediction.
\autoref{sec:prelim} demonstrated that this explanation is invalid by
providing a counterexample instance classified by the model as
$\lit{amphibian}$.
Applying \autoref{alg:repair} leads to the following correct
explanation:
\frmeq{
  \quad &
  \begin{array}{ll}
    \tn{\bf IF} &
    \nlit{feathers}\land\nlit{milk}\land\lit{backbone}\land\nlit{fins}\land\eqlit{legs}{\tn{0}}\land\lit{tail} \\[2pt]
    \tn{\bf THEN} & \eqlit{class}{reptile} \\
  \end{array}
  &
}
Note that although this explanation is larger than the one of Anchor,
it is global for the \emph{entire instance space}, i.e.\ there
guaranteed to be no counterexample for this explanation.
Also observe that \autoref{alg:repair} is able to keep features
$\nlit{milk}$ and $\nlit{fins}$ in the explanation even though there
\emph{may be} a repair with a fewer number of changes.

\SetKwBlock{Begin}{\texttt{begin}}{\texttt{end}}
\SetKw{Return}{\texttt{return}}
\SetKwIF{If}{ElseIf}{Else}{\texttt{if}}{\texttt{:}}{\texttt{elif}}{\texttt{else:}}{}%
\SetKwFor{ForEach}{\texttt{for}}{\texttt{:}}{}%
\SetKwFor{While}{\texttt{while}}{\texttt{:}}{}%

\SetKw{KwNot}{\texttt{not}\xspace}
\SetKw{KwAnd}{and\xspace}
\SetKw{KwOr}{or\xspace}
\SetKw{KwBreak}{break\xspace}
\SetKwData{false}{False}
\SetKwData{true}{True}
\SetKwData{st}{{\slshape st}}
\SetKwData{cores}{$\mathcal{C}$}
\SetKwFunction{ent}{Entails}
\SetKwFunction{ip}{IP}
\SetKwBlock{Let}{let}{end}
\SetKwBlock{FBlock}{}{end}

\SetKwFunction{SAT}{SAT}
\SetKwFunction{CNF}{CNF}
\SetKwFunction{minhs}{MinimumHS}
\SetKwFunction{model}{GetAssignment}
\SetKwFunction{falselits}{PickFalseLits}
\SetKwFunction{extractvals}{ExtractValues}

\begin{algorithm}[!t]
  \caption{Repairing an incorrect heuristic explanation} \label{alg:repair}

  \DontPrintSemicolon
  \SetAlgoNoLine
  \LinesNumbered
  \SetFillComment

  \KwIn{\,\,\,\,formula $\fml{M}$, input $I$, prediction $\pi$, and explanation $I_h$}
  \KwOut{correct explanation $I_m$}
  \BlankLine
  $I_{m_1}\gets I\setminus I_h$\;
  $I_{m_2}\gets I_h$\;
  \ForEach{$f \in I_{m_1}$}{
    \If{$\ent(I_{m_1}\cup I_{m_2}\setminus\{f\},\fml{M}\rightarrow\pi)$}{
      $I_{m_1}\gets I_{m_1}\setminus\{f\}$
    }
  }
  \ForEach{$f \in I_{m_2}$}{
    \If{$\ent(I_{m_1}\cup I_{m_2}\setminus\{f\},\fml{M}\rightarrow\pi)$}{
      $I_{m_2}\gets I_{m_2}\setminus\{f\}$
    }
  }
  \Return $I_{m_1}\cup I_{m_2}$ \;
  \BlankLine
\end{algorithm}


%% file: reduc.tex

\subsection{Refining Heuristic Explanations} \label{sec:reduc}

Yet another way to use XPlainer is to reduce a given explanation $I_h$
if it is proved by \autoref{alg:valid} to be valid.
Depending on the requirements of a user, this can be achieved by
applying either \autoref{alg:smexpl} or~\cite[Algorithm~2]{inms-aaai19}.
Here, the algorithms should receive the explanation $I_h\subseteq I$
instead of complete $I$.
As a result, they will output a subset- or cardinality-minimal
explanation $I_m$ s.t.\ $I_m\subseteq I_h$.

Although local explanations computed by heuristic approaches are
rarely globally correct, this approach is deemed a promising way to
prove the explanations to be minimal or to refine them further.
Note that due to the complexity of the abduction-based explanation
procedures, minimization of a given explanation may be significantly
more efficient than starting from a complete data instance: for
computing both subset- and cardinality-minimal explanations.


%% file: res.tex

\section{Experiments} \label{sec:res}

This section details the experimental results aiming at the assessment
of LIME and Anchor, the state-of-the-art heuristic approaches to
explaining black-box models.
Following \autoref{sec:cases}, it focuses
on validating, repairing, and refining heuristic explanations.

\subsection{Datasets and implementation} \label{sec:setup}
The results are obtained on the five well-known and publicly available
datasets.
Three of them were studied in~\cite{guestrin-aaai18} to illustrate the
advantage of Anchor's explanations over those of LIME including
\emph{adult}, \emph{lending}, and \emph{recidivism}.
These datasets were processed the same
way\footnote{\url{https://github.com/marcotcr/anchor-experiments}} as
in~\cite{guestrin-aaai18}.
The \emph{adult} dataset~\cite{kohavi-kdd96} is originally taken from
the Census bureau and targets predicting whether or not a given adult
person earns more than \$50K a year depending on various attributes.
%
The \emph{lending} dataset aims at predicting whether or not a loan on
the Lending Club website will turn out bad.
The \emph{recidivism} dataset was used to predict recidivism for
individuals released from North Carolina prisons in 1978 and
1980~\cite{schmidt-1988}.
Also, two additional datasets were considered including \emph{compas}
and \emph{german} that were previously studied in the context of the
FairML and Algorithmic Fairness projects~\cite{fairml17,fairness15}.
%
\emph{Compas} represents a popular dataset,
known~\cite{propublica16} for exhibiting racial bias of the
COMPAS algorithm used for scoring criminal defendant's
likelihood of reoffending.
The latter dataset is a German credit data (e.g.\
see~\cite{feldman-kdd15}), which given a list of people's attributes
classifies them as good or bad credit risks.

A prototype of \emph{XPlainer} (including the proposed encoding of
boosted trees and the explanation procedures) is implemented in
Python\footnote{XPlainer is available online:
\url{https://github.com/alexeyignatiev/xplainer}}.
Extraction of subset- and cardinali\-ty-minimal explanation follows
\autoref{alg:smexpl} and~\cite[Algorithm~2]{inms-aaai19},
respectively.
%
%
%
%
%
XPla\-iner makes use of SMT solver Z3~\cite{demoura-tacas08} as an
underlying reasoning engine.
%

%
%
%

\subsection{Results} \label{sec:results}
The performed experiment is detailed below.
First, following the standard setup, given a dataset, each XGBoost
model was trained on 80\% randomly chosen data instances.
Each XGBoost model contained 50 trees per class, each tree having
depth 3.
(Further increasing the number of trees per class and also increasing
the maximum depth of a tree does not result in a significant increase
of the models' accuracy on the training and test sets for the
considered datasets.)
Second, given a dataset and the trained model, an explanation for each
of the unique data instances\footnote{Datasets normally contain
  duplicate instances. Moreover, various predictions can be specified
  in the dataset for different instantiations of the same input. As
long as the classifier is trained, it behaves the same way for each of
the duplicates. As a result and in order to avoid unnecessary
repetition, each unique instance was considered once.} was computed
using either LIME\footnote{LIME expects to receive a target size of an
  explanation provided as input. Hence, the experiment bootstrapped
  LIME with the size of an \emph{existing} subset-minimal explanation
computed by XPlainer.} or Anchor.
Third, each explanation was then validated by XPlainer (see
\autoref{sec:valid}).
If an explanation was proved to be incorrect, i.e.\ \emph{optimistic},
XPlainer made an attempt to heuristically repair the explanation (see
\autoref{sec:fix}).
Otherwise, Xplainer tried to refine the explanation further (see
\autoref{sec:reduc}).
If succeeded, the explanation was treated as \emph{pessimistic}.
Otherwise, the explanation was reported to be correct and
subset-minimal, i.e.\ \emph{realistic} from the global perspective.

\begin{table}[!t]
  \caption{Heuristic explanations validated by XPlainer, for each
    unique data input of the considered datasets. The table shows the
    percentage of optimistic, pessimistic, and realistic explanations
    provided by LIME and Anchor. The total number of unique instances
  used is shown in column 2.}
  \label{tab:res}
  \begin{adjustbox}{center}
    \setlength{\tabcolsep}{0.5em}
    \begin{tabular}{ccS[table-format=2.1]S[table-format=2.1]S[table-format=2.1]S[table-format=2.1]S[table-format=2.1]S[table-format=2.1]}
      \toprule
      & & \multicolumn{6}{c}{\textbf{Explanations}} \\
      \cmidrule(lr){3-8} \textbf{Dataset} & \textbf{(\# unique)} &
      \multicolumn{2}{c}{\textbf{optimistic}} &
      \multicolumn{2}{c}{\textbf{pessimistic}} &
      \multicolumn{2}{c}{\textbf{realistic}} \\ \cmidrule(lr){3-4}
      \cmidrule(lr){5-6} \cmidrule(lr){7-8} & & \textbf{LIME} &
      \textbf{Anchor} & \textbf{LIME} & \textbf{Anchor} &
      \textbf{LIME} & \textbf{Anchor} \\
      \midrule
      %
      %
      adult & (5579) & 61.3\% & 80.5\% & 7.9\% & 1.6\% & 30.8\% & 17.9\% \\
      lending & (4414) & 24.0\% & 3.0\% & 0.4\% & 0.0\% & 75.6\% & 97.0\% \\
      recidivism & (3696) & 94.1\% & 99.4\% & 4.6\% & 0.4\% & 1.3\% & 0.2\% \\
      compas & (778) & 71.9\% & 84.4\% & 20.6\% & 1.7\% & 7.5\% & 13.9\% \\
      german & (1000) & 85.3\% & 99.7\% & 14.6\% & 0.2\% & 0.1\% & 0.1\% \\
      \bottomrule
    \end{tabular}
  \end{adjustbox}
\end{table}

The results of this experiment are shown in \autoref{tab:res}.
Although Anchor is supposed to improve over
LIME~\cite{guestrin-aaai18}, surprisingly, there is no clear winner
between LIME and Anchor; most explanations computed by either approach
are inadequate.
Observe that for the 4 out of 5 datasets the explanations of both LIME
and Anchor are mostly optimistic.
Concretely, for \emph{recidivism} and \emph{german} more than 99\% of
Anchor's explanations are optimistic.
Similar results hold for LIME, i.e.\ 94.1\% and 85.3\% explanations
for \emph{recidivism} and \emph{german} are optimistic.
The quality of Anchor's explanations improves for \emph{adult} and
\emph{compas} where there are more than 80\% of optimistic
explanations.
LIME is ahead of Anchor with 61.3\% and 71.9\% explanations being
optimistic for \emph{adult} and \emph{compas}.
Surprisingly, the result for the \emph{lending} dataset does not agree
with the rest, where only 3\% (24\%, resp.) of inputs are explained
incorrectly by Anchor (LIME, resp.).
Overall, 80.5\%, 3.0\%, 99.4\%, 84.4\%, and 99.7\% of Anchor's
explanations and 61.3\%, 24.0\%, 94.1\%, 71.9\%, 85.3\% of LIME's
explanations are optimistic (computed for \emph{adult},
\emph{lending}, \emph{recidivism}, \emph{compas}, and \emph{german},
respectively).
Also note that the number of pessimistic explanations is significantly
lower for Anchor.
Usually, there are less than 1.7\% of explanations that can be further
refined.
However, LIME can produce a significant number of them, e.g.\ for
\emph{adult}, \emph{compas}, and \emph{german} the percentage of
pessimistic explanations reaches 7.9\%, 20.6\%, and 14.6\%,
respectively.

Explanations for the remaining data instances were proved to be
correct and subset-minimal (see column marked by \emph{realistic}).
For Anchor, these comprise 17.9\%, 97.0\%, 0.2\%, 13.9\%, 0.1\% of
inputs for \emph{adult}, \emph{lending}, \emph{recidivism},
\emph{compas}, and \emph{german}, respectively.
For LIME, the percentage of realistic explanations is 30.8\%, 75.6\%,
1.3\%, 7.5\%, and 0.1\% for \emph{adult}, \emph{lending},
\emph{recidivism}, \emph{compas}, and \emph{german}, respectively.

\begin{table}[!t]
  \caption{Contribution of Anchor and XPlainer to the average total
  runtime (in seconds). \emph{Repair}, \emph{refinement}, and,
\emph{validation} time is denoted by `Rep.', `Ref.', and `Valid.',
respectively.}
  \label{tab:time}
  \begin{adjustbox}{center}
    \setlength{\tabcolsep}{0.56em}
    \begin{tabular}{cccccccc}
      %
      \toprule
      \multirow{2}{*}{\textbf{Dataset}} &
      \multirow{2}{*}{\textbf{LIME}} &
      \multirow{2}{*}{\textbf{Anchor}} &
      \multirow{2}{*}{\textbf{Valid.}} &
      \multicolumn{2}{c}{\textbf{Subset-minimal}} &
      \multicolumn{2}{c}{\textbf{Cardinality-minimal}} \\
      \cmidrule(lr){5-6} \cmidrule(lr){7-8} & & & & \textbf{Rep.} &
      \textbf{Ref.} & \textbf{Rep.}\tablefootnote{Note that in the
      case of cardinality-minimal mode of XPlainer, \emph{repair}
    essentially means computing a smallest size explanation \emph{from
  scratch}.} & \textbf{Ref.} \\
      \midrule
      adult & 0.43 & 0.20 & 0.17 & 0.41 & 0.33 & 1.58 & 0.54 \\
      lending & 0.28 & 0.10 & 0.05 & 0.23 & 0.07 & 0.37 & 0.10 \\
      rcdv & 0.52 & 0.29 & 0.11 & 0.29 & 0.31 & 1.18 & 0.42 \\
      compas & 0.37 & 0.18 & 0.06 & 0.12 & 0.09 & 0.23 & 0.11 \\
      german & 0.64 & 0.39 & 1.33 & 8.26 & 1.89 & 63.02 & 7.14 \\
      \bottomrule
    \end{tabular}
  \end{adjustbox}
\end{table}

To conclude, there are cases when Anchor and LIME behave reasonably
well, e.g.\ for the \emph{lending} dataset.
However, as the \autoref{tab:res} indicates, in most situations the
explanations provided by both heuristic approaches are either globally
incorrect (optimistic) or can be further refined (pessimistic).

Contribution of LIME, Anchor, and XPlainer (including the
\emph{validation}, \emph{repair}, and \emph{refinement} time) to the
average total runtime for each data instance is shown in
\autoref{tab:time}.
Observe that validation time is usually negligible.
Also, repairing and refining heuristic explanations in XPlainer's
subset-minimal mode is consistently faster than in the
cardinality-minimal mode.
This is especially the case for the \emph{german} dataset, which is
the hardest for XPlainer to deal with.

Now, let us compare the size of explanations produced by Anchor and
XPlainer\footnote{LIME is not shown here as LIME's explanations and
the subset-minimal ones are equal in size.}.
\autoref{tab:size} details the comparison showing the minimum,
maximum, and mean values, as well as standard deviation for the
explanations computed by Anchor and also subset- and
cardinality-minimal explanations computed by XPlainer.
Here, given an explanation of Anchor, XPlainer was instructed to
either repair or refine it.
It is not surprising that the mean value for the size of Anchor's
explanations is typically lower, because, as was shown above, most of
the time Anchor's explanations are globally optimistic.
In general, the average size of Anchor's explanations varies from 10\%
to 23\% of the total number of features.
The average size of subset-minimal explanations is 15--50\%, which is
still quite good in terms of interpretability.
Furthermore, cardinality-minimal explanations improve this result to
15--36\% of features on average.
However and as the experimental results confirm (see
\autoref{tab:time}), computing a cardinality-minimal explanation is
computationally more expensive.
This represents a reasonable trade-off: depending on user's
requirements, XPlainer can be applied to compute a subset-minimal
explanation (faster but worse quality) or to compute a
cardinality-minimal explanation (slower but better quality).

\begin{table}[!t]
  \caption{Explanation size of Anchor vs XPlainer. Here, `m',
    `M', `$\mu$', and $\sigma$ denote a minimum, maximum, mean values,
    and a standard deviation, respectively. Column~1
    shows the name of a dataset followed by the total number of
  features.}
  \begin{adjustbox}{center}
  \small
  \setlength{\tabcolsep}{0.75em}
  \begin{tabular}{cccccccccccccc}
    \toprule
    \multirow{2}{*}{\textbf{Dataset}} & &
    \multicolumn{4}{c}{\textbf{Anchor}} &
    \multicolumn{4}{c}{\textbf{Subset-minimal}} &
    \multicolumn{4}{c}{\textbf{Cardinality-minimal}} \\
    \cmidrule(lr){3-6}
    \cmidrule(lr){7-10}
    \cmidrule(lr){11-14}
    & & m & M & $\mu$ & $\sigma$ & m & M & $\mu$ & $\sigma$ & m & M &
    $\mu$ & $\sigma$ \\
    \midrule
    adult & {\scriptsize (12)} & 1 & 11 & 2.8 & 1.3 & 1 & 11 & 4.2 &
    1.4 & 1 & 11 & 3.9 & 1.3 \\
    lending & {\scriptsize (9)} & 1 & 5 & 1.3 & 0.7 & 1 & 6 & 1.4 &
    0.8 & 1 & 6 & 1.4 & 0.8 \\
    rcvd & {\scriptsize (15)} & 2 & 14 & 3.3 & 1.3 & 4 & 11 & 6.0 &
    1.3 & 4 & 11 & 5.5 & 1.2 \\
    compas & {\scriptsize (11)} & 1 & 11 & 2.2 & 1.4 & 2 & 9 & 4.4 &
    1.5 & 2 & 9 & 3.6 & 1.0 \\
    german & {\scriptsize (21)} & 1 & 21 & 2.2 & 2.5 & 5 & 15 & 9.0 &
    1.8 & 3 & 15 & 6.3 & 2.5 \\
    \bottomrule
  \end{tabular}
  \end{adjustbox}
  \label{tab:size}
\end{table}


%% file: relw.tex

\section{Related Work} \label{sec:relw}

\anoteF{1 page for related work + conclusions}

The importance of providing explanations for predictions made by ML
models has grown in significance in recent years, motivated both by
ongoing research programs~\cite{darpa-xai16}, but also by recently
approved legislation~\cite{eu-reg16,goodman-aimag17}. Nevertheless,
the importance of explanations can be traced until the mid
90s~\cite{shavlik-nips95}.
%
%
%
The computation of explanations can be broadly organized into two main
categories, depending on whether the ML model considered is
interpretable or not. An ML model is viewed as interpretable if it is
amenable to interpretation by a human decision maker. This is the case
with decision trees, lists or sets. When considering interpretable ML
models, the goal is then to compute models that provide \emph{minimal}
explanations associated with each prediction. A number of works has
addressed this topic
recently~\cite{leskovec-kdd16,rudin-kdd17,nipms-ijcai18,ipnms-ijcar18}.
The work on generating explainable (interpretable) models can be
further organized into heuristic approaches
(e.g.\ \cite{leskovec-kdd16}) and exact
solutions~\cite{rudin-kdd17,nipms-ijcai18,ipnms-ijcar18}.
Clearly, a limitation of these approaches is that they are restricted to
interpretable ML models, which in many settings are not the preferred
choice.
A different alternative consists of (heuristically) \emph{compiling}
a non-interpretable ML model into another (interpretable)
one~\cite{hinton-cexaiia17}, but an assessment from a (global) quality
viewpoint is unavailable.
Recent compilation-based approaches for computing global explanations
consider Bayesian network classifiers~\cite{darwiche-ijcai18}, with
the drawback of exponential worst-case compilation sizes.
For non-interpretable models, one line of work is based on sensitivity
analysis~\cite{muller-jmlr10,muller-dsp18}, on the use of simulated
annealing~\cite{gales-asru15}, or on the use of
case-based reasoning~\cite{rudin-aaai18}. Recent methods attempt to
improve interpretability of non-interpretable models by analysis of
the model after training. Recent work reached conclusions similar to
ours with respect to saliency methods~\cite{kim-nips18a}.
With few exceptions, existing
approaches~\cite{kononenko-dke09,kononenko-jmlr10,muller-jmlr10,doshi-velez-nips15,gales-asru15,doshi-velez-ijcai17,muller-dsp18,rudin-aaai18,doshi-velez-aaai18a,doshi-velez-aaai18b,darwiche-ijcai18,kim-nips18a,jaakkola-nips18}
are local in nature, and although some are efficient in practice,
computed explanations offer no global guarantees similar to the ones
provided by XPlainer.
Exact compilation approaches~\cite{darwiche-ijcai18} are one such
exception, but also exhibit similar (if not worse) concerns in terms
of scalability.

\jnoteF{ToDo:
  \begin{enumerate}
  \item Include subsection on related
    work. Cover~\cite{shavlik-nips95,kononenko-dke09,kononenko-jmlr10,muller-jmlr10,rudin-jmlr17,muller-dsp18,rudin-aaai18}.
  \item Also cover~\cite{leskovec-kdd16,rudin-icdm16,rudin-jmlr17,rudin-kdd17,rudin-aaai18}.
  \end{enumerate}
}


%% file: conc.tex

\section{Conclusions} \label{sec:conc}

This paper extends earlier work on computing provably correct
explanations, by considering the concrete case of boosted
trees~\cite{guestrin-kdd16a}.
The proposed encoding is shown to scale to realistic sized boosted
trees, either for computing subset-minimal and cardinali\-ty-minimal
(correct, global) explanations.
In turn, this enabled a first assessment of recently proposed
heuristic approaches for computing
explanations~\cite{guestrin-kdd16b,guestrin-aaai18}. On the datasets
considered, the results are conclusive and indicate that existing
heuristic approaches may be either too optimistic, thus, overlooking
feature values that are necessary to provide a global explanation of a
prediction, or pessimistic, i.e.\ containing a number of redundant
feature values.

A possible downside of the proposed approach is scalability. The
NP-hardness of finding subset-minimal (and the $\stwop$-hardness
of finding cardinality-minimal~\cite{umans-tcad06}) explanations is
likely to limit the applicability of XPlainer. Nevertheless, as the
results demonstrate, XPlainer is well-suited to assess the quality of
existing and new heuristic approaches, on small to medium-scale ML
models.

Given the experimental results in this paper, one line of work
is to devise more robust heuristic approaches for explaining
non-interpretable ML models. Another line of work is to assess other
heuristic approaches for explaining ML models,
e.g.\ \cite{muller-jmlr10,lee-nips17,rudin-aaai18}.
Although not a concern for the ML models studied in this paper, a
third line of work is to improve the underlying reasoning engine(s)
and the proposed encodings, aiming at better scalability of the
(provably correct) explanations obtained with XPlainer on more complex
ML models.
One final line of work is to extend XPlainer to other
non-interpretable ML models.
